\documentclass[sigconf,screen]{acmart}
\AtBeginDocument{%
  }

\setcopyright{acmlicensed}
\copyrightyear{2026}
\acmYear{2026}
\acmDOI{XXXXXXX.XXXXXXX}
\acmConference[Conference]{Make sure to enter the correct
  conference title from your rights confirmation email}{2026}{ }
\acmISBN{978-1-4503-XXXX-X/2018/06}

\renewcommand\footnotetextcopyrightpermission[1]{} 
\acmSubmissionID{6766}

\usepackage{multirow}
\usepackage{threeparttable}

\begin{document}

\title{RT-NeRV: Rethinking Hybrid Neural Representations for Video via Residual Tokenization}

\author{Yunjie Xu}
\email{xuyunjie13@gmail.com}
\affiliation{%
  \institution{Ningbo University}
  \city{Ningbo}
  \country{China}
}

\author{Xiang Feng}
\email{xiangfeng@hdu.edu.cn}
\affiliation{%
  \institution{Hangzhou Dianzi University}
  \city{Hangzhou}
  \country{China}}

\author{Chengkai Wang}
\email{williamwang0128@gmail.com}
\affiliation{%
  \institution{Hangzhou Dianzi University}
  \city{Hangzhou}
  \country{China}}

\author{Alan Wee-Chung Liew}
\email{a.liew@griffith.edu.au}
\affiliation{%
 \institution{Griffith University}
 \city{Gold Coast}
 \country{Australia}}

\author{Xuefei Yin}
\email{x.yin@griffith.edu.au}
\affiliation{%
 \institution{Griffith University}
 \city{Gold Coast}
 \country{Australia}}

 \author{Yanming Zhu}
\email{yanming.zhu@griffith.edu.au}
\affiliation{%
 \institution{Griffith University}
 \city{Gold Coast}
 \country{Australia}}

\renewcommand{\shortauthors}{Xu et al.}

\begin{abstract}
Neural Representations for Videos (NeRV) have emerged as a promising paradigm for video compression by representing videos as compact neural networks with efficient decoding. Hybrid NeRV methods further improve reconstruction quality through content-adaptive embeddings, but still struggle to preserve fine details at low bitrates. A key limitation is that shallow residual support information, although highly beneficial for reconstruction, is costly to transmit in its continuous form and is therefore underutilized. In this paper, we rethink hybrid NeRV and present \textbf{RT-NeRV}, a residual tokenization framework for hybrid neural video representations. The core idea is to discretize shallow residual features and inter-frame residual cues into compact residual tokens, allowing informative reconstruction support to be transmitted efficiently and exploited by the decoder. To this end, we design a residual tokenizer together with a residual-aware codebook learning strategy that improves token utilization and stabilizes training. RT-NeRV can be readily integrated into modern hybrid NeRV hosts, consistently enhancing detail preservation, reconstruction quality, and bitrate-quality trade-offs. Extensive experiments on video regression and related restoration tasks show that RT-NeRV outperforms strong hybrid NeRV baselines and remains competitive with recent INR-based video compression methods. These results demonstrate that residual tokenization is an effective and complementary direction for advancing hybrid neural video representations. 
\end{abstract}

\ccsdesc[500]{Computing methodologies~Computer vision}


\keywords{video compression, implicit neural representations, hybrid NeRV, residual tokenization, vector quantization, detail preservation}


\maketitle


\section{Introduction}
\label{sec:intro}
Recent advances in neural video compression have shown that videos can be represented as compact neural models~\cite{agustsson2020scale,habibian2019video} rather than only by hand-crafted codecs~\cite{wiegand2003overview,sullivan2012overview}. Among them, Implicit Neural Representations (INRs), especially Neural Representations for Videos, NeRV~\cite{chen2021nerv}, provide an appealing paradigm by representing a video as a compact neural network with efficient feed-forward decoding. To overcome the limited representation capacity of content-agnostic embeddings, HNeRV~\cite{chen2023hnerv} further introduces a hybrid architecture with content-adaptive embeddings, significantly improving reconstruction quality. Despite this progress, preserving fine structures under low bitrate budgets remains a fundamental challenge for hybrid NeRV methods.

A key limitation is the inadequate use of reconstruction-supportive shallow information. In encoder-decoder architectures, shallow features are important for preserving high-frequency details, yet directly transmitting them in compression is prohibitively expensive. As a result, existing hybrid NeRV methods largely discard this information to reduce bitrate, which weakens decoder-side reconstruction and often causes detail loss in fine textures and other high-frequency regions. This raises a key question: how can shallow residual support information be preserved compactly and effectively exploited for neural video reconstruction?

In this paper, we revisit hybrid NeRV from the perspective of \emph{residual tokenization} and propose \textbf{RT-NeRV}, a residual tokenization framework for hybrid neural video representations. The key idea is to convert shallow residual features and inter-frame residual cues from costly continuous features into compact discrete tokens that preserve useful reconstruction support. RT-NeRV introduces a residual tokenizer to map shallow residual support information into a finite token space, allowing the decoder to recover skip-like cues with low bitrate overhead. We further develop a residual-aware codebook optimization strategy to improve token utilization and training stability. As a result, RT-NeRV improves detail preservation and bitrate-quality trade-offs while maintaining a compact hybrid neural representation. Extensive experiments on video regression and restoration tasks show that RT-NeRV consistently improves reconstruction fidelity over strong hybrid NeRV baselines and remains competitive with recent INR-based video compression methods, demonstrating that residual tokenization is a practical and complementary direction for advancing hybrid NeRV.

The main contributions of this work are summarized as follows:
\begin{itemize}
\item We revisit hybrid neural video representations through residual tokenization, identifying shallow residual support information as a critical yet underutilized source for detail-preserving reconstruction under bitrate constraints.

\item We propose \textbf{RT-NeRV}, a residual tokenization framework that discretizes shallow residual features and inter-frame residual cues into compact residual tokens, enabling efficient transmission of reconstruction-supportive information for decoder-side enhancement.

\item We develop a residual-aware codebook optimization strategy that improves token utilization and training stability, yielding stronger reconstruction quality and better bitrate-quality trade-offs.

\item We demonstrate that residual tokenization is effective and complementary to recent advances in INR-based video compression through extensive experiments on video regression and related restoration tasks.
\end{itemize}

\begin{figure*}[t] 
    \centering 
    \includegraphics[width=\linewidth]{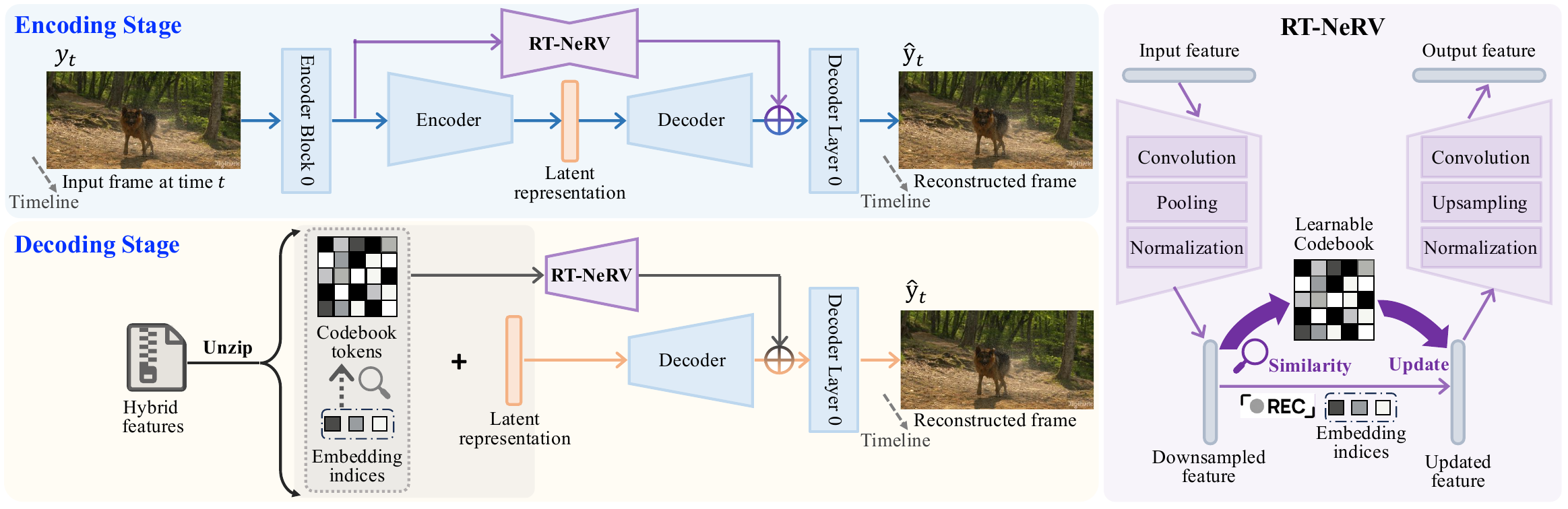} 
    \vspace{-0.5cm}
    \caption{Overview of RT-NeRV. Built on a hybrid NeRV host, RT-NeRV inserts a residual tokenizer between the first encoder block and the last decoder layer to convert shallow reconstruction-supportive features into compact codebook tokens. The transmitted token indices and codebook are used at decoding time to reconstruct residual support for decoder-side enhancement.} 
    \label{fig:framework} 
\end{figure*}

\section{Related Work}
\label{sec:relate}
\subsection{NeRV for Video Compression}
Implicit neural representations (INRs) model signals as neural functions and have been widely studied for image reconstruction~\cite{sitzmann2020implicit}, shape representation~\cite{park2019deepsdf}, and 3D scene synthesis~\cite{feng2023zs,feng2024srgs,feng2026sr3r}. For video compression, Neural Representations for Videos (NeRV)~\cite{chen2021nerv} introduced a frame-wise formulation that maps frame indices directly to full video frames, reformulating video compression as model compression. Subsequent works, such as E-NeRV~\cite{li2022nerv} and DNeRV~\cite{zhao2023dnerv}, further improved video regression quality. These methods establish the basic NeRV paradigm, but improving reconstruction fidelity under limited bitrate budgets remains a central challenge.

\subsection{Hybrid NeRV for Fine-grained Reconstruction}
To improve the limited representation capacity of content-agnostic embeddings, HNeRV~\cite{chen2023hnerv} introduced a hybrid neural representation with content-adaptive embeddings, significantly improving reconstruction quality. Subsequent hybrid designs further enhanced decoder conditioning and feature utilization, demonstrating the effectiveness of content-aware latent representations for neural video compression. However, existing hybrid NeRV methods still rely primarily on compact latent embeddings while largely discarding shallow reconstruction-supportive information. In encoder-decoder architectures, shallow features are crucial for preserving high-frequency details and local structures, yet directly transmitting them is prohibitively expensive in a compression setting. As a result, hybrid NeRV methods often suffer from weakened detail preservation, especially in fine textures and high-frequency regions. Our work is motivated by this unresolved tension between reconstruction support and transmission efficiency.

\subsection{Recent Advances Beyond Hybrid NeRV}
Recent NeRV-based methods have advanced the NeRV family from several complementary directions. Tree-NeRV~\cite{zhao2025tree}, RNeRV~\cite{gwilliam2026design}, TeNeRV~\cite{zhu2026video}, TeCoNeRV~\cite{padmanabhan2026teconerv}, and SRNeRV~\cite{wang2026srnerv} improve NeRV through non-uniform temporal sampling, training-time-aware design, temporal modeling, encoding efficiency, and parameter sharing, while FaNeRV/FANeRV~\cite{zhu2026frequency, li2025fanerv} and NVRC~\cite{kwan2024nvrc} further strengthen spectral modeling and compression framework design. These methods substantially broaden the NeRV landscape. However, they mainly improve temporal modeling, training efficiency, parameter sharing, spectral modeling, or framework-level compression, and do not explicitly investigate how shallow residual support information can be compactly tokenized and transmitted for detail-preserving reconstruction in hybrid NeRV. Our work complements these directions by focusing on residual tokenization as an efficient mechanism for decoder-side reconstruction support.

\subsection{Discrete Tokenization and Compact Reconstruction Support}
Discrete representation learning, especially vector quantization, provides an effective way to convert continuous features into compact tokenized representations. Representative works such as VQ-VAE~\cite{van2017neural} show that codebooks can offer efficient discrete latent spaces while maintaining strong reconstruction capability. This property is particularly attractive for video compression, since discrete tokens are more compact and easier to encode than dense continuous features. Motivated by this idea, our work revisits hybrid neural video representations from the perspective of residual tokenization, aiming to preserve shallow residual support information in a bitrate-efficient form.

\section{Method}
\subsection{Framework Overview}
\label{sec:overview}

We propose \textbf{RT-NeRV}, a residual tokenization framework for hybrid NeRV. RT-NeRV is designed for encoder-decoder hybrid hosts and aims to preserve shallow reconstruction-supportive information in a compact discrete form. As shown in Fig.~\ref{fig:framework}, RT-NeRV is inserted between the first encoder block and the last decoder layer of a hybrid NeRV host. Given an input frame $y_t$, the encoder first extracts a shallow feature from its early stage, while the host network also produces a compact latent representation for the main decoding path. Instead of directly forwarding the shallow feature, RT-NeRV transforms it into a compact tokenized residual representation. Specifically, the input feature is first downsampled by a learnable projection network to a lower-resolution feature space, reducing the representation size and enabling a more bitrate-efficient codebook. The resulting feature map is then flattened into a set of spatial descriptors. For each descriptor, RT-NeRV retrieves the nearest entry from a learnable codebook and substitutes the original feature vector with the selected token. After tokenization over all spatial locations, the feature is fully represented by codebook tokens and their corresponding indices. The tokenized feature is then mapped back to the original feature resolution through a learnable upsampling network, producing an output feature with the same spatial size as the original shallow feature.

Since the output of RT-NeRV matches the size of the inserted shallow feature, it can be naturally integrated into the encoder-decoder pipeline. The recovered residual feature is concatenated with the decoder feature before the final decoding layer, which enables the decoder to exploit additional reconstruction-supportive information for detail-preserving frame reconstruction. In this way, RT-NeRV acts as a compact residual bridge between the encoder and decoder, providing skip-like support without explicitly transmitting high-dimensional shallow features.

After encoding a video, the transmitted bitstream contains four components: the per-frame latent representations from the hybrid host, the learned codebook, the codebook token indices, and the trained decoder parameters. During decoding, the latent representations are fed into the decoder, while the codebook and token indices are fed into RT-NeRV. Notably, only the reconstruction branch of RT-NeRV, namely the upsampling side, is required at the decoder. RT-NeRV first reconstructs the tokenized residual feature from the received codebook tokens, and this feature is then injected into the decoder to reconstruct the compressed video frame.

To further reduce bitrate, we exploit temporal redundancy in token indices. Instead of transmitting complete token indices for every frame, we transmit the full token indices only for the first frame, and encode subsequent frames using index residuals with respect to the previous frame. Since adjacent video frames usually exhibit small feature variations, these index residuals are considerably more compact than transmitting full indices for every frame. This design further improves the bitrate efficiency of RT-NeRV while preserving decoder-side reconstruction support.

\subsection{Residual Tokenizer}
\label{sec:tokenizer}
The core component of RT-NeRV is a \emph{residual tokenizer}, which converts shallow reconstruction-supportive features into a compact discrete representation. Given a hybrid NeRV host, we take the shallow feature extracted from the first encoder block of frame $y_t$ as the input to RT-NeRV, denoted by
\begin{align}
    f_t^{e} \in \mathbb{R}^{H \times W \times C},
\end{align}
where $H$, $W$, and $C$ denote the spatial height, spatial width, and channel dimension, respectively.

\noindent \textbf{Learnable downsampling.}
Directly transmitting $f_t^{e}$ is prohibitively expensive. We therefore first map it into a lower-resolution latent feature space using a learnable downsampling network:
\begin{align}
    z_t = \mathcal{D}(f_t^{e}), \qquad
    z_t \in \mathbb{R}^{h \times w \times C},
\end{align}
where $\mathcal{D}(\cdot)$ denotes the learnable downsampling function, and $h \ll H$, $w \ll W$. This step reduces the spatial redundancy of the shallow feature and makes it possible to learn a more bitrate-efficient codebook in a compact feature space.

\noindent \textbf{Spatial tokenization via codebook matching.}
We then reshape the downsampled feature into a set of spatial descriptors:
\begin{align}
    \bar{z}_t = \mathrm{Reshape}(z_t) \in \mathbb{R}^{N \times C}, \qquad N = hw,
\end{align}
where each row $\bar{z}_{t,i} \in \mathbb{R}^{C}$ corresponds to the feature vector at one spatial location. We define a learnable codebook
\begin{align}
    \mathcal{Z} = \{e_k\}_{k=1}^{K}, \qquad e_k \in \mathbb{R}^{C},
\end{align}
where $K$ is the number of codebook entries. For each spatial descriptor $\bar{z}_{t,i}$, RT-NeRV searches for its nearest codeword by
\begin{align}
    q_{t,i} = \arg\min_{k \in \{1,\dots,K\}} \|\bar{z}_{t,i} - e_k\|_2^2,
\end{align}
where $q_{t,i}$ is the token index assigned to the $i$-th spatial location.

\noindent \textbf{Token substitution.}
After nearest-neighbor search, each spatial descriptor is replaced by its matched codeword:
\begin{align}
    \tilde{z}_{t,i} = e_{q_{t,i}}.
\end{align}
Applying this substitution to all spatial locations yields the fully tokenized feature
\begin{align}
    \tilde{z}_t = \mathrm{Reshape}^{-1}([\tilde{z}_{t,1}, \tilde{z}_{t,2}, \dots, \tilde{z}_{t,N}])
    \in \mathbb{R}^{h \times w \times C}.
\end{align}
Therefore, the original continuous shallow feature is converted into a compact discrete representation described by the codebook and the token index set
\begin{align}
    q_t = \{q_{t,1}, q_{t,2}, \dots, q_{t,N}\}.
\end{align}

\noindent \textbf{Learnable upsampling and decoder-side injection.}
To make the tokenized feature compatible with the host decoder, RT-NeRV maps it back to the original feature resolution through a learnable upsampling network:
\begin{align}
    \hat{f}_t^{e} = \mathcal{U}(\tilde{z}_t), \qquad
    \hat{f}_t^{e} \in \mathbb{R}^{H' \times W' \times C'}.
\end{align}
In our implementation, $\hat{f}_t^{e}$ is designed to have the same size as the feature at the insertion point before the final decoder layer, so that it can be naturally integrated into the host network. Let $f_t^{d}$ denote the decoder feature before the last decoding layer. We concatenate the reconstructed residual feature with the decoder feature and perform the final reconstruction as
\begin{align}
    \hat{y}_t = \mathrm{Dec}_0\big(\mathrm{Concat}(f_t^{d}, \hat{f}_t^{e})\big),
\end{align}
where $\mathrm{Dec}_0(\cdot)$ denotes the last decoder layer.

Through this design, RT-NeRV acts as a compact residual bridge between the encoder and decoder. Instead of transmitting dense shallow features, it communicates only the learned codebook and token indices, while still providing the decoder with effective reconstruction-supportive information for preserving fine details. During decoding, only the reconstruction branch of RT-NeRV is required. Specifically, the received codebook and token indices are first used to recover the tokenized feature $\tilde{z}_t$, which is then mapped by the upsampling network $\mathcal{U}(\cdot)$ to produce $\hat{f}_t^{e}$ for decoder-side enhancement.

\subsection{Residual-aware Codebook Learning}
\label{sec:codebook}
A practical challenge in residual tokenization is low codebook utilization. Since RT-NeRV quantizes shallow reconstruction-supportive descriptors into a finite set of codewords, the effectiveness of the residual tokenizer depends heavily on whether the learned codebook can adequately cover the residual feature space. In practice, however, only a small subset of codewords may be frequently activated, while many others remain rarely used or even inactive. This problem is particularly severe in our setting because hybrid NeRV training typically uses a very small batch size, making codebook updates based only on the current mini-batch unstable.

To address this issue, we introduce a \emph{residual-aware codebook learning} strategy. Instead of updating the codebook solely from the current batch, we maintain a feature pool that stores representative residual descriptors observed during training. Let
\begin{align}
    \mathcal{P} = \{p_m\}_{m=1}^{M}, \qquad p_m \in \mathbb{R}^{C},
\end{align}
denote the feature pool, where each entry is a residual descriptor collected from the downsampled feature space. This pool provides a more stable approximation of the global residual feature distribution than the current mini-batch alone.

During training, each spatial descriptor $\bar z_{t,i}$ is first assigned to its nearest codeword by
\begin{align}
    q_{t,i} = \arg\min_{k \in \{1,\dots,K\}} \|\bar z_{t,i} - e_k\|_2^2 .
\end{align}
Based on the resulting assignments, we update codeword statistics using an exponential moving average strategy:
\begin{align}
    n_k \leftarrow \gamma n_k + (1-\gamma)\sum_{t,i}\mathbf{1}[q_{t,i}=k],
\end{align}
\begin{align}
    m_k \leftarrow \gamma m_k + (1-\gamma)\sum_{t,i}\mathbf{1}[q_{t,i}=k]\bar z_{t,i},
\end{align}
\begin{align}
    e_k \leftarrow \frac{m_k}{n_k + \epsilon},
\end{align}
where $n_k$ and $m_k$ denote the running count and accumulated feature mean of the $k$-th codeword, respectively, $\gamma$ is the EMA momentum, and $\epsilon$ is a small constant for numerical stability.

To further improve utilization, we periodically refresh underused codewords. If the usage count of a codeword falls below a threshold $\tau$, we treat it as inactive and replace it with a representative residual descriptor from the feature pool. In practice, we first cluster the feature pool to obtain stable residual prototypes, and then reassign each inactive codeword to the nearest sufficiently used prototype. This strategy is more reliable than random replacement under small-batch training and helps prevent codebook collapse.

This residual-aware codebook learning strategy offers two benefits: it improves token utilization and stabilizes training under the low-batch regime common in hybrid NeRV optimization. As a result, RT-NeRV preserves more informative reconstruction-supportive residual cues with a compact token budget, leading to better bitrate-quality trade-offs.

\subsection{Training Objective and Bitstream Formation}
\label{sec:objective}
RT-NeRV is trained end-to-end together with the hybrid NeRV host. The training objective consists of a reconstruction term and a quantization term, which jointly encourage high-fidelity frame reconstruction and compact residual tokenization.

\noindent \textbf{Reconstruction loss.}
Let $y_t$ and $\hat{y}_t$ denote the ground-truth frame and the reconstructed frame at time $t$, respectively. To preserve both pixel fidelity and structural consistency, we adopt a hybrid reconstruction loss:
\begin{align}
\mathcal{L}_{rec}
=
\alpha \| \hat{y}_t - y_t \|_1
+
(1-\alpha)\bigl(1-\mathrm{SSIM}(\hat{y}_t, y_t)\bigr),
\end{align}
where $\alpha=0.7$ controls the balance between the $\ell_1$ and the SSIM. 

\noindent \textbf{Quantization loss.}
Let $z_t \in \mathbb{R}^{h \times w \times C}$ be the downsampled feature produced by the residual tokenizer, and let $\tilde{z}_t$ be its tokenized counterpart obtained from the codebook. We use the standard vector quantization objective
\begin{align}
\mathcal{L}_{q}
=
\left\| \mathrm{sg}[z_t] - \tilde{z}_t \right\|_2^2
+
\beta \left\| z_t - \mathrm{sg}[\tilde{z}_t] \right\|_2^2,
\end{align}
where $\mathrm{sg}[\cdot]$ denotes the stop-gradient operator, and $\beta$ is the commitment weight. The first term updates the codebook toward the residual descriptors, while the second term encourages the tokenizer output to stay close to its assigned codewords.

\noindent \textbf{Overall objective.}
The final training objective is
\begin{align}
\mathcal{L}_{total} = \mathcal{L}_{rec} + \lambda_q \mathcal{L}_{q},
\end{align}
where $\lambda_q$ balances reconstruction quality and quantization regularization.

\noindent \textbf{Bitstream formation.}
After training on a target video, RT-NeRV forms the compressed representation using four components: (1) the latent representations produced by the hybrid NeRV host, (2) the learned codebook $\mathcal{Z}$, (3) the token indices generated by the residual tokenizer, and (4) the decoder-side parameters used for reconstruction. More specifically, for the first frame we transmit the complete token index map $q_1$. For subsequent frames, instead of transmitting the full token index map $q_t$, we transmit only its residual with respect to the previous frame,
\begin{align}
\Delta q_t = q_t - q_{t-1}, \qquad t \geq 2.
\end{align}
This design exploits temporal redundancy in videos, since adjacent frames usually produce similar token assignments. As a result, the index residuals are substantially more compact than transmitting full token maps for every frame.

Let $B_{lat}$, $B_{code}$, $B_{idx}$, and $B_{dec}$ denote the bit costs of the latent representations, codebook, token indices, and decoder-side parameters, respectively. The total bitstream size can be written as
\begin{align}
B_{total} = B_{lat} + B_{code} + B_{idx} + B_{dec},
\end{align}
with
\begin{align}
B_{idx} = B(q_1) + \sum_{t=2}^{T} B(\Delta q_t),
\end{align}
where $T$ is the number of frames, $B(q_1)$ denotes the cost of the first-frame token index map, and $B(\Delta q_t)$ denotes the cost of the index residual at frame $t$. In practice, the latent representations, codebook, and index stream can be further compressed by a lossless entropy coder. During decoding, the latent representations are fed into the host decoder, while the codebook and the received token indices are used by RT-NeRV to reconstruct the residual supportive feature for decoder-side enhancement.

\section{Experiments}
\subsection{Datasets, Settings, and Evaluation Protocol}
\label{sec:exp_setup}

\noindent \textbf{Datasets.}
We evaluate RT-NeRV on three commonly used benchmarks: Big Buck Bunny (Bunny)~\cite{roosendaal2008big}, UVG~\cite{mercat2020uvg}, and DAVIS~\cite{wang2016mcl}. Bunny contains 132 frames at a resolution of $720\times1280$. UVG contains 7 videos at $1080\times1920$ with high frame rates and durations of 2.5s or 5s. DAVIS is a high-quality benchmark widely used for video analysis and restoration. Following prior hybrid NeRV settings \cite{chen2023hnerv}, we center-crop the videos to $640\times1280$ or $960\times1920$ depending on the experiment.

\noindent \textbf{Host backbones.}
To validate the generality of residual tokenization on hybrid NeRVs, we instantiate RT-NeRV on two host backbones: HNeRV \cite{chen2023hnerv} and HNeRV-Boost \cite{zhang2024boosting}. 

\noindent \textbf{Implementation details.}
We use the Adam optimizer with $\beta_1=0.9$ and $\beta_2=0.999$, zero weight decay, and an initial learning rate of $10^{-3}$ with cosine decay. The batch size is set to 1. For the residual tokenizer, the codebook size is chosen according to the input resolution, following the same configuration protocol as our previous implementation. Unless otherwise stated, the codebook size is set to 1024 for $1280\times640$ inputs and 2048 for $1920\times960$ inputs, and the dead-code threshold is set to 1. Other architectural hyperparameters follow the corresponding host backbone settings.

\noindent \textbf{Evaluation metrics.}
We evaluate reconstruction quality using PSNR and MS-SSIM. Compression efficiency is measured by bits-per-pixel (bpp), and we report rate-distortion curves whenever sufficient points are available. In addition, we report efficiency-related metrics including encoding time, decoding frames per second (FPS), and multiply-accumulate operations (MACs). 

\noindent \textbf{Comparison protocol.}
We compare RT-NeRV with representative NeRV-based methods, including NeRV~\cite{chen2021nerv}, E-NeRV~\cite{li2022nerv}, HNeRV~\cite{chen2023hnerv}, PS-NeRV~\cite{bai2023ps}, DNeRV~\cite{zhao2023dnerv}, D-NeRV~\cite{he2023towards}, FFNeRV~\cite{lee2023ffnerv}, and HiNeRV~\cite{kwan2024hinerv}, as well as recent INR-based video compression methods, including CANeRV~\cite{tang2025canerv}, FANeRV~\cite{li2025fanerv}, FaNeRV~\cite{zhu2026frequency}, MSNeRV~\cite{zhu2025msnerv}, NVRC~\cite{kwan2024nvrc}, SRNeRV~\cite{wang2026srnerv}, and Tree-NeRV~\cite{zhao2025tree}. HNeRV, HNeRV-Boost, and their RT-NeRV-enhanced variants are evaluated under a unified implementation setting. Since official code is unavailable for PS-NeRV and the recent methods above, we report the results provided in their original papers under matched datasets and resolutions whenever possible. If only partial metrics are reported, we include only the available quantities and mark the others as unavailable. For methods that provide only PSNR--bpp curves, we use them for qualitative rate-distortion comparison without deriving additional metrics. Encoding time, decoding FPS, and MACs are reported only when directly measured by us or explicitly available in the original papers.

\noindent \textbf{Task settings.}
For video regression, we follow the standard full-sequence reconstruction protocol. For video interpolation, we train on the even-frame sequence and test on the odd-frame sequence. For video inpainting, we directly evaluate the trained regression models on masked videos without additional fine-tuning, following the same protocol as in our earlier implementation.

\begin{table*}[t]
    \caption{Host-level comparison of video regression on the UVG dataset with a matched 3M decoder budget. RT-NeRV is instantiated on two hybrid NeRV hosts, HNeRV and HNeRV-Boost. We report reconstruction quality and computational cost under the same training schedule.}
    \vspace{-0.2cm}
    \centering
    \resizebox{\linewidth}{!}
    {
    \begin{tabular}{l cccccccccc cccc}
    \toprule[1pt]
     \multirow{2}{*}{Method} & \multicolumn{8}{c}{PSNR $\uparrow$} & \multirow{2}{*}{MACs $\downarrow$} & \multirow{2}{*}{Encoding Time $\downarrow$} & \multirow{2}{*}{Decoding FPS $\uparrow$} \\
     \cmidrule(lr){2-9}
      & Beauty & Bospho & Honey & Jockey & Ready & Shake & Yach & Avg & &  & \\
    \midrule[1pt]
    HNeRV~\cite{chen2023hnerv} & 33.58 & 34.73 & 38.96 & 32.04 & 25.74 & 34.57 & 29.26 &32.69 & \textbf{28.84g} & \textbf{24.57ms} &  \textbf{225}\\
    RT-NeRV(HNeRV)  & \textbf{34.15} & \textbf{36.89} & \textbf{39.41} & \textbf{36.77} & \textbf{31.74} & \textbf{36.08} & \textbf{32.85} & \textbf{35.41} & 36.05g & 34.89ms & 160\\
    \midrule
    HNeRV-Boost~\cite{zhang2024boosting} & 33.80 & 36.11 & 39.65 & 34.28 & 28.19 & 35.88 & 29.33 &  33.89 & \textbf{118.02g} & \textbf{181.37ms} & \textbf{43}  \\
    RT-NeRV(Boost) & \textbf{34.42}  & \textbf{39.33}  & \textbf{39.66} & \textbf{37.37} & \textbf{32.26} & \textbf{36.40}  & \textbf{33.53} & \textbf{36.14} &  141.01g & 256.42ms & 31 \\
    \bottomrule[1pt]
    \end{tabular}
    }
    \label{tab:UVG_main}
\end{table*}

\begin{table*}[t]
    \centering
    \caption{Robustness across model capacities and training epochs on hybrid NeRV hosts. PSNR is reported for different decoder sizes and epochs at a 1.5M decoder on the Bunny dataset to assess the stability of RT-NeRV across capacity and training settings.}
    \vspace{-0.2cm}
    \small
    \begin{tabular}{l@{\hspace{12pt}} ccccc@{\hspace{8pt}} cccccc}
    \toprule[1pt]
    \multirow{2}{*}{Method} & \multicolumn{4}{c}{Size} && \multicolumn{6}{c}{Epoch} \\
    \cmidrule(lr){2-5} \cmidrule(lr){7-12}
    & 0.35M & 0.75M & 1.5M & 3M && 50 & 100 & 150 & 200 & 250 & 300\\
    \midrule[1pt]
    HNeRV~\cite{chen2023hnerv}    & 30.15 & 32.81 &35.57 &37.43 && 26.95 & 30.74 & 33.03 & 34.93 & 35.32 & 35.59 \\
    RT-NeRV(HNeRV)  & \textbf{31.06} & \textbf{34.00} & \textbf{36.61} & \textbf{39.05}  && \textbf{27.01} &  \textbf{32.72} &  \textbf{34.76} &  \textbf{35.89} &  \textbf{36.61} &  \textbf{36.61} \\
    \midrule
    HNeRV-Boost~\cite{zhang2024boosting} & 34.20 & 36.21 & 37.81 & 41.85  && 26.75 & 34.12 & 35.77 & 36.89 & 37.71 & 37.85\\
    RT-NeRV(Boost)  &  \textbf{34.45} &  \textbf{36.72} &  \textbf{38.68} &  \textbf{42.05} &&  \textbf{31.69} &  \textbf{34.38} &  \textbf{36.73} &  \textbf{37.74} &  \textbf{38.56} &  \textbf{38.68} \\
    \bottomrule[1pt]
    \end{tabular}
    \label{tab:Bunny_main}
\end{table*}

\begin{figure*}[t]
\centering
\includegraphics[width=\linewidth]{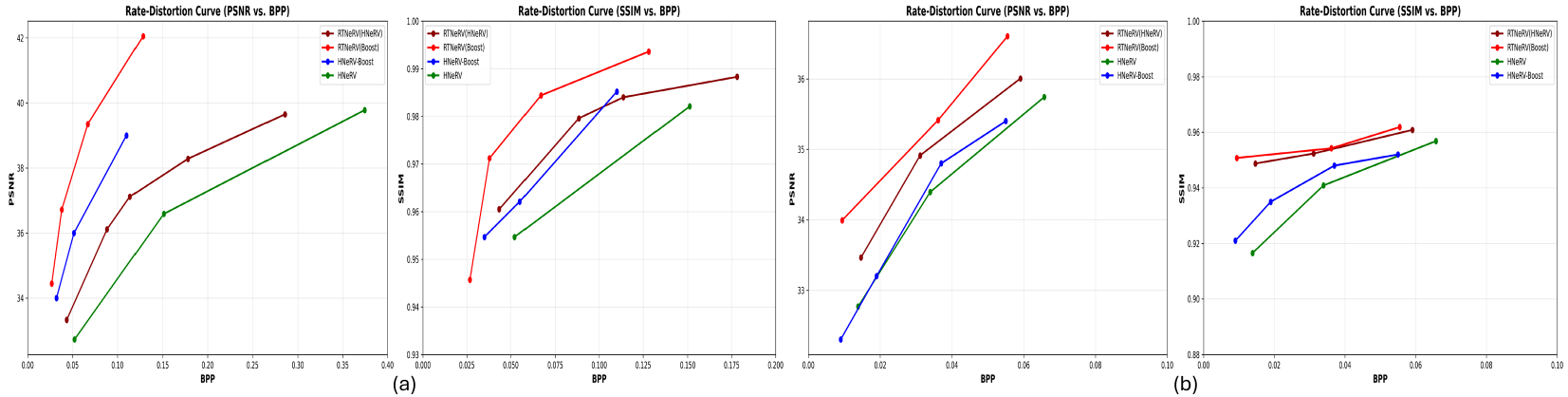}
\vspace{-0.6cm}
\caption{Rate-distortion curves on (a) Bunny and (b) UVG datasets. RT-NeRV instantiated on HNeRV and HNeRV-Boost consistently improves the bitrate-quality trade-off in terms of both PSNR and MS-SSIM.}
\label{fig:curve_main}
\end{figure*}

\subsection{Host-level Validation on Hybrid NeRV Backbones}
\label{sec:host_validation}
We first evaluate whether residual tokenization consistently improves hybrid NeRVs. To this end, we compare two representative backbones, HNeRV and HNeRV-Boost, with and without RT-NeRV under the same training schedules.

Table~\ref{tab:UVG_main} reports the host-level comparison on the UVG dataset at a representative operating point. RT-NeRV consistently improves reconstruction quality on both hosts, yielding clear gains in average PSNR as well as per-sequence performance. These results show that the benefit of residual tokenization is not limited to a single backbone, but generalizes across different hybrid NeRV hosts. In particular, RT-NeRV brings stronger improvements on challenging sequences with richer structures and motion variations, which suggests that the proposed residual tokenizer effectively provides reconstruction-supportive information for preserving fine details.

We also report computational metrics including MACs, encoding time, and decoding FPS. As expected, introducing RT-NeRV incurs additional computation compared with the original hosts. However, the added cost remains moderate relative to the significant reconstruction gains, indicating a favorable accuracy-efficiency trade-off. This supports our design goal: rather than introducing a heavy auxiliary branch, RT-NeRV provides a compact residual bridge that enhances decoder-side reconstruction with limited overhead.

To further examine robustness, Table~\ref{tab:Bunny_main} compares the host baselines and their RT-NeRV-enhanced variants across different model capacities and training epochs on the Bunny dataset. RT-NeRV consistently improves the host backbones over a wide range of decoder sizes, showing that the proposed residual tokenization is effective not only at a single operating point but also across different capacity regimes. The same trend is observed under different training epochs, where RT-NeRV maintains a clear advantage throughout training. These results confirm that the gains of RT-NeRV are stable and not tied to a specific model scale or training budget.

The bitrate-quality trade-off is further illustrated by the rate-distortion curves in Fig.~\ref{fig:curve_main}. On both datasets, RT-NeRV shifts the host curves toward a better operating region, demonstrating that the proposed residual tokenization improves reconstruction quality under comparable bitrate budgets. Overall, these results verify that RT-NeRV provides a robust and transferable improvement for hybrid NeRV backbones.

\begin{figure*}[t] 
    \centering 
    \includegraphics[width=\linewidth]{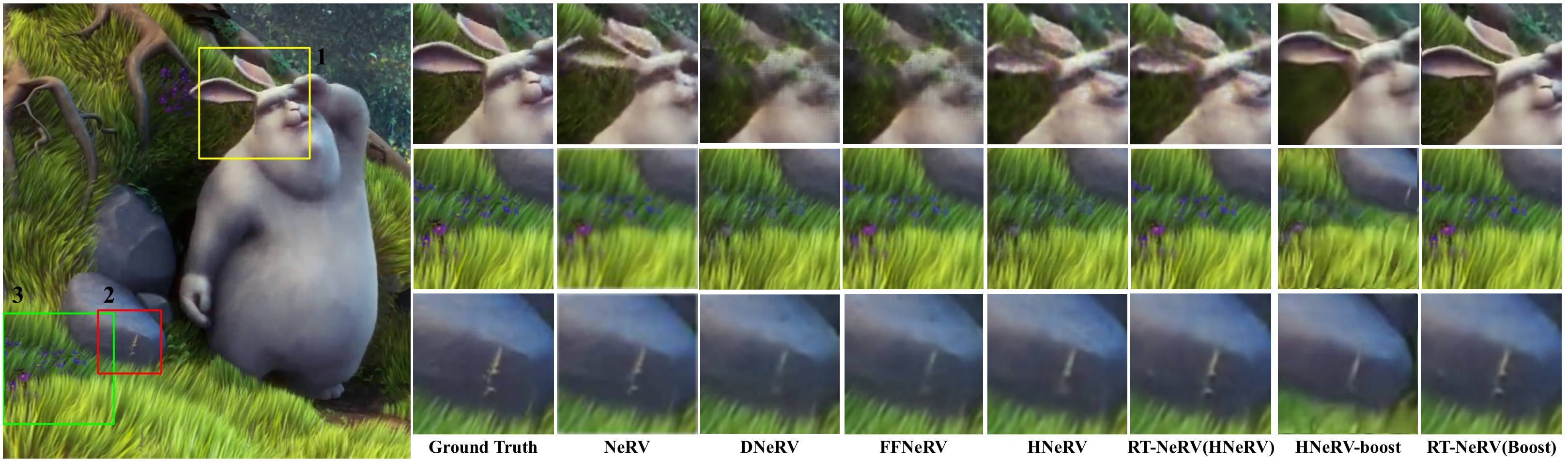} 
    \vspace{-0.5cm}
    \caption{Qualitative comparison on the 131st frame of the Bunny dataset under the same extremely low decoder budget (0.35M). RT-NeRV consistently preserves clearer boundaries, richer local textures, and more faithful fine-grained structures than representative NeRV-based baselines.}
    \label{fig:visual} 
\end{figure*}

\subsection{Comparison with Representative NeRV-based Methods}
\label{sec:main_comparison}
We further compare RT-NeRV with representative NeRV-based video representation methods under a matched decoder budget. As shown in Table~\ref{tab:all}, RT-NeRV consistently achieves the best overall reconstruction quality among all compared methods, yielding the highest average PSNR and the strongest or highly competitive performance on each individual sequence. These results demonstrate that the proposed residual tokenization provides a substantial advantage over both early implicit video representations and more recent NeRV-family variants.

A notable observation is that RT-NeRV maintains clear gains not only over earlier baselines such as NeRV, E-NeRV, and PS-NeRV, but also over stronger hybrid and recent NeRV-based methods. This suggests that the benefit of RT-NeRV does not come from a simple increase in model capacity, but from a more effective use of reconstruction-supportive residual information. By explicitly tokenizing shallow residual cues and injecting them into the decoder in a compact form, RT-NeRV strengthens fine-grained reconstruction while preserving the compactness of the host representation.

\begin{figure}[t] 
    \centering 
    \includegraphics[width=0.65\linewidth]{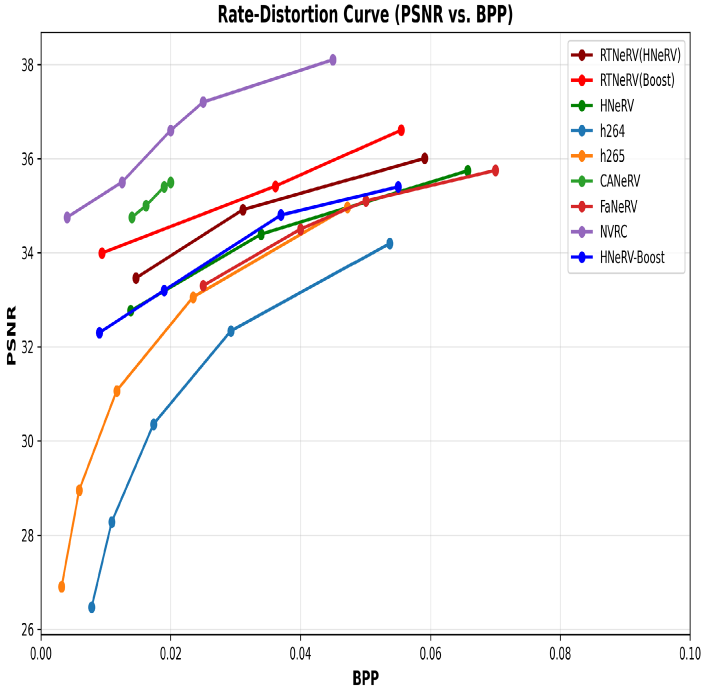} 
    \vspace{-0.4cm}
    \caption{Rate-distortion comparison on the UVG dataset.}
    \vspace{-0.2cm}
    \label{fig:rd_recent} 
\end{figure}

Fig.~\ref{fig:rd_recent} further compares RT-NeRV with recent INR-based video compression methods through the rate-distortion curves. Since several recent methods do not release public code, and some report only partial metrics or only PSNR--bpp curves, we use the quantities explicitly reported in their original papers under matched settings whenever possible. Despite this limitation, RT-NeRV remains competitive with the strongest recent methods and outperforms several recent NeRV-family baselines in the bitrate-quality trade-off.

\subsection{Detail Preservation and Visual Quality}
\label{sec:visual_quality}

To further evaluate the visual effect of RT-NeRV, we compare it with representative NeRV-based methods on the Bunny dataset under an extremely low decoder budget. As shown in Fig.~\ref{fig:visual_compare}, all methods are evaluated at the same decoder size, making the visual comparison directly comparable. RT-NeRV consistently produces clearer structures and more faithful local textures than the competing methods. In the selected patches, previous methods often suffer from over-smoothing, blurred boundaries, and loss of fine-grained details, especially in high-frequency regions such as the ear contour, grass texture, and thin object structures. In contrast, RT-NeRV preserves clearer structural transitions and richer texture patterns. These improvements are consistently observed when RT-NeRV is instantiated on both HNeRV and HNeRV-Boost, indicating that the proposed residual tokenization effectively enhances decoder-side reconstruction by providing compact reconstruction-supportive information. These qualitative results show that RT-NeRV is particularly effective in low-bitrate regimes, where preserving reconstruction-supportive shallow information becomes critical. By converting such information into compact residual tokens and reinjecting them into the decoder, RT-NeRV achieves stronger detail preservation and more faithful reconstruction than existing baselines.

\begin{table}[t]
    \caption{Comparison with representative NeRV-based methods on video regression at resolution $960\times1920$ on the UVG dataset under a matched 3M decoder setting. All methods follow the same experimental setting whenever reproducible. For methods without public code, we report the results from the original papers under matched dataset and resolution settings.}
    \vspace{-0.2cm}
    \centering
    \small
    \setlength{\tabcolsep}{1.5pt}
    \resizebox{\linewidth}{!}
    {
    \begin{tabular}{l cccccccccc}
    \toprule[1pt]
     \multirow{2}{*}{Method} & \multicolumn{8}{c}{PSNR $\uparrow$} \\
     \cmidrule(lr){2-9}
      & Beauty & Bospho & Honey & Jockey & Ready & Shake & Yach & Avg \\
    \midrule[1pt]
        NeRV~\cite{chen2021nerv} & 33.25 & 33.22 & 37.25 & 31.74 & 24.84 & 33.08 & 28.03 & 31.63 \\
        E-NeRV~\cite{li2022nerv} & 33.17 & 33.69 & 37.63 & 31.63 & 25.24 & 34.39 & 28.42 & 32.02 \\
        PS-NeRV~\cite{bai2023ps} & 32.94 & 32.32 & 38.39 & 30.38 & 23.61 & 33.26 & 26.33 & 31.13 \\
        DNeRV~\cite{zhao2023dnerv} & 33.91 & 33.17 & 38.22 & 32.01 & 25.56 & 33.58 & 27.15 & 31.94 \\
        D-NeRV~\cite{he2023towards} & 33.77 & {38.66} & 37.97 & 35.51 & {33.93} & 35.04 & \textbf{33.73} & {35.52} \\
        FFNeRV~\cite{lee2023ffnerv} & 33.57 & 35.03 & 38.95 & 31.57 & 25.92 & 34.41 & 28.99 & 32.63 \\
        HiNeRV~\cite{kwan2024hinerv} & 34.08 & 36.57 & {39.71} & 36.10 & 31.53 & 35.85 & 30.95 & 34.97 \\
        FANeRV~\cite{li2025fanerv} & 33.93 & 37.09 & 39.69 & 35.36 & 29.68 & 35.97 & 30.58 & 34.64 \\
        MSNeRV~\cite{zhu2025msnerv} & 34.17 & 38.95 & \textbf{39.71} & \textbf{37.37} & \textbf{32.59} & 35.85 & 31.55 & 34.70 \\
        RT-NeRV(Boost) & \textbf{34.42}  & \textbf{39.33}  & {39.66} & \textbf{37.37} & {32.26} & \textbf{36.40}  & {33.53} & \textbf{36.14} \\
    \bottomrule[1pt]
    \end{tabular}
    }
    \label{tab:all}
\end{table}

\begin{table*}[t]
        \caption{Video inpainting results (PSNR$\uparrow$) on the DAVIS dataset with five fixed box masks. RT-NeRV is evaluated on the HNeRV host and consistently improves masked-region reconstruction quality.}
        \vspace{-0.2cm}
        \centering
        \begin{tabular}{l cccccccccc c}
            \toprule
            Method  & bike & b-swan & bmx & b-danc & camel & c-roun & c-shado & cows & d-twirl & dog & Avg\\
            \midrule
            Input    & 23.14 & 20.24 & 19.99 & 21.36 & 17.3 & 20.47 & 18.92 & 19.37 & 20.45 & 18.39 & 19.96 \\
            HNeRV~\cite{chen2023hnerv} & 28.37 & 29.49 & 29.77 & 27.86 & 27.4 & 26.94 & 29.96 & 24.15 & 27.89 & 30.72 & 28.26 \\
            RT-NeRV(HNeRV) & \textbf{31.74} & \textbf{32.26} & \textbf{33.22} & \textbf{31.12} & \textbf{27.4} & \textbf{28.53} & \textbf{33.57} & \textbf{24.89} & \textbf{28.26} & \textbf{33.38} & \textbf{30.44} \\
            \bottomrule
        \end{tabular}
\\
 \footnotesize{``Input'' denotes the masked input video evaluated directly against the ground truth.}
        \label{tab:inpainting}
\end{table*}

\begin{table*}[t]
     \caption{Video interpolation results (PSNR$\uparrow$) on the UVG and DAVIS datasets. RT-NeRV is evaluated on the HNeRV host and shows improved generalization to unseen intermediate frames.}
    \vspace{-0.2cm}
    \centering
    \begin{tabular}{l cccccccccc c}
        \toprule
        Method  & beauty & b-swan & bmx & b-danc & camel & bee & jockey & ready & shake & yach & Avg\\
        \midrule
        HNeRV~\cite{chen2023hnerv} & 31.10 & \textbf{21.44} & 18.29 & 20.29 & 20.64 & 38.83 & 23.82 & 20.99 & 32.61 & \textbf{27.24} & 25.52 \\
        RT-NeRV(HNeRV) & \textbf{31.15} & 21.08 & \textbf{19.33} & \textbf{20.63} & \textbf{20.64} & \textbf{38.98} & \textbf{24.33} & \textbf{21.18} & \textbf{32.75} & 27.15 & \textbf{25.72} \\
        \bottomrule
    \end{tabular}
    \label{tab:interpolation}
\end{table*}

\subsection{Downstream Restoration Tasks}
\label{sec:downstream}
We further evaluate RT-NeRV on downstream restoration tasks, including video inpainting and video interpolation. We report the restoration results on the HNeRV host here to assess whether residual tokenization also benefits representation quality beyond standard video regression.

\noindent \textbf{Video Inpainting.}
We first evaluate RT-NeRV on fixed-mask video inpainting. Following the setting of HNeRV, we use five fixed box masks with width 50 on DAVIS videos and directly test the trained models without additional task-specific redesign. Since the official testing scripts of HNeRV on DAVIS are unavailable, we re-evaluate HNeRV under the same setting for a fair comparison. As shown in Table~\ref{tab:inpainting}, RT-NeRV consistently outperforms the HNeRV baseline across most sequences and achieves a clear improvement in average PSNR. These results indicate that the proposed residual tokenization provides more effective reconstruction-supportive information when part of the visual content is missing. In particular, by preserving compact shallow residual cues, RT-NeRV helps the decoder recover sharper local structures and more faithful textures in masked regions.

\noindent \textbf{Video Interpolation.}
We also evaluate video interpolation to assess the representation generalization ability of RT-NeRV. Following the standard protocol, we train on the even frames and test on the odd frames on UVG and DAVIS. The quantitative results are reported in Table~\ref{tab:interpolation}. Video interpolation remains challenging for INR-based representations, since these models are optimized toward fitting the observed training frames. Nevertheless, RT-NeRV still achieves a modest overall improvement over HNeRV and obtains better results on several sequences with more noticeable motion variations. This suggests that residual tokenization not only improves reconstruction fidelity on observed frames, but also helps retain more informative local cues that benefit interpolation.

Overall, the downstream restoration results provide additional evidence that RT-NeRV improves the quality of hybrid NeRVs beyond the primary compression task. The gains are especially clear for video inpainting, where recovering missing local content relies more heavily on effective reconstruction support, which is precisely the target of our residual tokenization design.

\subsection{Ablation Study}
\label{sec:ablation}
We perform ablation studies on the Bunny dataset to better understand the contribution of the key components in RT-NeRV. 

\noindent \textbf{Contribution of the residual tokenizer and residual-aware codebook learning.}

The first three rows of Table~\ref{tab:ablation} evaluate the impact of the two core components of RT-NeRV. Compared with the HNeRV baseline, adding the residual tokenizer yields a clear PSNR improvement, showing that compactly tokenizing shallow reconstruction-supportive information is beneficial for video reconstruction. When the residual-aware codebook learning strategy is further introduced, the performance improves again, together with a substantial increase in codebook utilization. This result confirms that the gain of RT-NeRV comes not only from introducing discrete residual tokens, but also from learning a well-utilized codebook that effectively represents the residual feature space.

\noindent \textbf{Effect of different codebook settings.}
Table~\ref{tab:ablation} further studies the influence of codebook design choices. First, Euclidean matching performs better than cosine matching, indicating that distance-based nearest-neighbor assignment is more suitable for the residual tokenization setting. Second, increasing the codebook dimension generally improves reconstruction quality, suggesting that richer token representations better preserve reconstruction-supportive information. Third, smaller codebooks tend to yield higher usage, but overly small codebooks reduce representation capacity and lead to inferior reconstruction quality. In contrast, overly large codebooks improve flexibility but suffer from low utilization. These observations reveal a clear trade-off between codebook expressiveness and usage efficiency, which motivates our final configuration.

Overall, the ablation results validate the main design of RT-NeRV. The residual tokenizer provides strong reconstruction gains through compact residual support, and the residual-aware codebook learning strategy further improves this benefit through better codebook utilization and training stability. Together, these results explain the consistent gains of RT-NeRV in the main experiments.

\begin{table}[t]
    \caption{Ablation of RT-NeRV on the Bunny dataset. We report PSNR and codebook usage on the HNeRV host to analyze the contribution of the residual tokenizer, residual-aware codebook learning strategy, and different codebook settings.}
    \vspace{-0.1cm}
    \centering
    \begin{tabular}{l cc}
    \toprule
    Method  & PSNR $\uparrow$ & Usage $\uparrow$ \\
    \midrule
    baseline                        & 35.52     & - \\
    +RT-NeRV Block                  & 36.25     & 39.26\%  \\
    +Shallow Codebook Optimization  & 36.60      & 100\% \\
    codebook\_dim=2                 & 36.20     & 37.70\% \\
    codebook\_dim=4                 & 36.54     & 37.70\% \\
    codebook\_dim=64                & 36.61     & 37.70\% \\
    use\_cosine                     & 35.82     & 26.37\% \\
    codebook\_size=128              & 36.19     & 88.28\% \\
    codebook\_size=256              & 36.58     & 64.45\% \\
    codebook\_size=1,024             & 36.60     & 22.27\% \\
    \bottomrule
    \end{tabular}
    \label{tab:ablation}
\end{table}

\section{Conclusion}
We revisited hybrid neural video representations from the perspective of residual tokenization and proposed RT-NeRV, a simple yet effective framework for compactly preserving shallow reconstruction-supportive information. By converting such residual cues into discrete tokens and reinjecting them into the decoder, RT-NeRV consistently improves detail preservation and bitrate-quality trade-offs on strong hybrid NeRV hosts. Extensive experiments further show that residual tokenization is an effective and complementary direction for advancing hybrid neural video representations.


\bibliographystyle{ACM-Reference-Format}
\bibliography{sample-base}



\end{document}